\documentclass[runningheads,a4paper]{llncs}
\usepackage{amssymb}
\usepackage{graphicx,epsfig,setspace,subfig,url,amsmath}
\usepackage{algorithm}
\usepackage[noend]{algpseudocode}
\usepackage{longtable}
\usepackage{array}
\usepackage{amsmath}
\usepackage{mathtools}
\usepackage{caption}
\usepackage{multirow}
\usepackage{relsize}

\usepackage[T1]{fontenc}
\usepackage[font=small,labelfont=bf,tableposition=top]{caption}

\def \hfillx {\hspace*{-\textwidth} \hfill}

\urldef{\mailsa}\path|{gabriel.maicas@adelaide.edu.au}|	
\urldef{\mailsb}\path|{gustavo.carneiro@adelaide.edu.au}|
\urldef{\mailsc}\path|{bradley@itee.uq.edu.au}|
\urldef{\mailsb}\path|{jan@isr.ist.utl.pt}|
\urldef{\mailsb}\path|{ian.reid@adelaide.edu.au}|
\newcommand{\keywords}[1]{\par\addvspace\baselineskip\noindent\keywordname\enspace\ignorespaces#1}
 \normalsize

\begin{document}
\title{Training Medical Image Analysis Systems like Radiologists \thanks{Supported by Australian Research Council through grants DP180103232, CE140100016 and FL130100102.}}
\institute {$^{\dagger}$Australian Institute for Machine Learning, The University of  Adelaide \\ $^{\ddagger}$Science and Engineering Faculty, Queensland University of Technology  \\ $^{\ddagger\ddagger}$Institute for Systems and Robotics, Instituto Superior Tecnico, Portugal}
\author{Gabriel Maicas$^{\dagger}$ \qquad  Andrew P. Bradley$^{\ddagger}$ \qquad Jacinto C. Nascimento$^{\ddagger\ddagger}$ \\ Ian Reid$^{\dagger}$ \qquad Gustavo Carneiro$^{\dagger}$}

\maketitle

\vspace{-.1in} 
\begin{abstract}

The training of medical image analysis systems using machine learning approaches follows a common script: collect and annotate a large dataset, train the classifier on the training set, and test it on a hold-out test set.  This process bears no direct resemblance with radiologist training, which is based on solving a series of tasks of increasing difficulty, where each task involves the use of significantly smaller datasets than those used in machine learning.  In this paper, we propose a novel training approach inspired by how radiologists are trained.  In particular, we explore the use of meta-training that models a classifier based on a series of tasks. Tasks are selected using teacher-student curriculum learning, where each task consists of simple classification problems containing small training sets.  We hypothesize that our proposed meta-training approach can be used to pre-train medical image analysis models.  This hypothesis is tested on the automatic breast screening classification from DCE-MRI trained with weakly labeled datasets.  The classification performance achieved by our approach is shown to be the best in the field for that application, compared to state of art baseline approaches: DenseNet, multiple instance learning and multi-task learning.

\vspace{-.2in} 
-\keywords{meta-learning, curriculum learning, multi-task training, breast image analysis, breast screening, magnetic resonance imaging.}
\end{abstract}

\vspace{-.1in} 
\section{Introduction}
\label{sec:intro}
\vspace{-.1in}
Radiologists are exceptionally trained specialists who play a crucial role interpreting and assisting other doctors and specialists in diagnosing and treating diseases.
Their training program typically requires the trainee to solve tasks of increasing difficulty~\cite{radTrain}, where each task contains a relatively small number of "training images".
Such a program bears little resemblance to the training of medical image analysis systems based on machine learning that are modeled to solve narrowly defined, but complex classification problems~\cite{wang2017chestx}, requiring large training sets.  
Once trained, these models cannot be easily adapted to new problems -- they must be re-trained with new large training sets.
The use of pre-trained models~\cite{bar2015deep} as a way of initializing a model is the first step towards a more similar approach to the training program of radiologists.  
However, pre-training does not train a model to be able to learn new tasks -- instead it is a "trick" to improve  convergence and generalization.  
Meanwhile, machine learning researchers have developed more effective \emph{learning to learn} approaches~\cite{finn2017model} -- such approaches are motivated by the ability of humans to learn new tasks quickly and with limited "training sets".  
The optimization in such approaches penalizes classification loss and inefficient learning on new tasks (i.e., classification problems) by using a training scheme that continuously samples new tasks, mimicking the human training process.
Our hypothesis is that medical machine learning methods could benefit from such an radiologist's style training process.

In this paper, we introduce an improved model agnostic meta-learning~\cite{finn2017model} (MAML) as a way of pre-training a classifier. 
The training process maximizes the ability of the classifier to adapt to new tasks using relatively small training sets.  We also propose a technical innovation for MAML~\cite{finn2017model}, by replacing the random task selection with teacher-student curriculum learning as an improved way for selecting tasks~\cite{matiisen2017teacher}.  This task selection process is based on the model's performance on the tasks, trying to mimic radiologists' training.  Our improved MAML is tested on weakly-supervised breast screening from DCE-MRI, where samples are globally annotated with classes (i.e. volume-level labels): \emph{no findings}, \emph{benign lesions} and \emph{malignant lesions}, but these samples do not have lesion delineations.  Note that the use of weakly-labeled datasets is becoming increasingly important for medical image analysis as this is the data available in clinical practice~\cite{wang2017chestx}.

We test our proposed approach on a dataset of dynamic contrast enhanced MRI for the breast screening classification. Results show that our proposed approach improves the 
area under the ROC curve (AUC), outperforming baselines such as DenseNet~\cite{huang2017densely}, which holds the state-of-the-art (SOTA)  for many classification problems; multiple-instance learning~\cite{zhu2017deep}, which holds SOTA for breast screening in mammography; 
and multi-task learning~\cite{xue2018full}.
Our learning approach produces an AUC of 0.90, which is better than the best result from the baseline methods that achieves an AUC of 0.85.

\vspace{-.05in} 
\section{Literature Review}
\label{sec:lit_review}
\vspace{-.05in} 

Breast screening from DCE-MRI aims at early detection of breast cancer in women at high-risk~\cite{smith2017cancer}.  Currently, this screening process is mostly done manually, where its success depends on the radiologist's abilities~\cite{vreemann2018frequency}.  
An automated breast screening system working as a second reader can help radiologists reduce variability and increase the sensitivity and specificity of their readings. Traditionally, such systems rely on classifiers trained with large-scale strongly labeled datasets (i.e., containing lesion delineation and global classification)~\cite{gubern2015automated,Ufuk2018Fully,amit2017hybrid,jager2017revealing,maicas2017deep}. 
The non-scalability of this process (due to costs related to the annotation process) motivated the development of learning methods that can use weakly-labeled training sets~\cite{zhu2017deep} (i.e., samples contain only global classification).  
However, these methods still follow traditional machine learning approaches, which means that they still need large-scale training sets, even when the model has been pre-trained from other classification problems~\cite{bar2015deep}.

Contrasting with traditional machine learning algorithms, humans excel at learning new skills and new "classification" problems, where new learning tasks often require fewer training samples than the ones before. 
This \emph{learning to learn} ability has inspired the development of a new generation of machine learning algorithms.  
For example, multi-task learning uses an optimization function that is trained to simultaneously minimize the loss of several different, but related classification problems~\cite{xue2018full}, helping the regularization of the training procedure.
Nevertheless, multi-task learning does not address the issue of making a model effective at learning new classification problems with small datasets.
This issue is addressed by \textit{meta-learning}~\cite{finn2017model}, which has been designed to solve the \textit{few-shot} learning problem, where the classifier is trained to train for new classification problems with previously unseen classes containing a small number of images.
In meta-learning for few-shot classification, the model is \emph{meta-trained} to solve classification problems for many randomly sampled tasks (i.e., the tasks are not fixed as in multi-task learning).  
Then the model is \emph{meta-tested} by classifying unseen classes after being able to adapt using \textit{few} training images of such unseen classes.

We explore the potential to improve the meta-learning process using a more useful (i.e., non random) task sampling procedure.
For example, formulating the task sampling as a multi-armed bandit problem has been shown to produce faster convergence and better generalization~\cite{gutierrez2017multi}. 
Similarly, Matiisen {\em et al.}~\cite{matiisen2017teacher}  proposed a new form of curriculum learning~\cite{bengio2009curriculum} that selects new tasks based not on their performance but on their performance improvement.
However, these task sampling approaches have been applied in traditional machine learning problems, such as supervised and reinforcement learning problems, 
which means that our proposed application of curriculum learning for task selection in meta-learning is novel, to the best of our knowledge~\footnote{While writing the final draft of this paper, we noticed a recent approach by Sharma et al.~\cite{sharma2018learning}. However, they sample tasks for the problem of multi-task learning. In addition, sampling tasks is not based on the improvement of performance, but tasks where the performance is worse.}.

\begin{figure}[t]
\begin{center}
\begin{tabular}{c}
\includegraphics[width=0.7\textwidth]{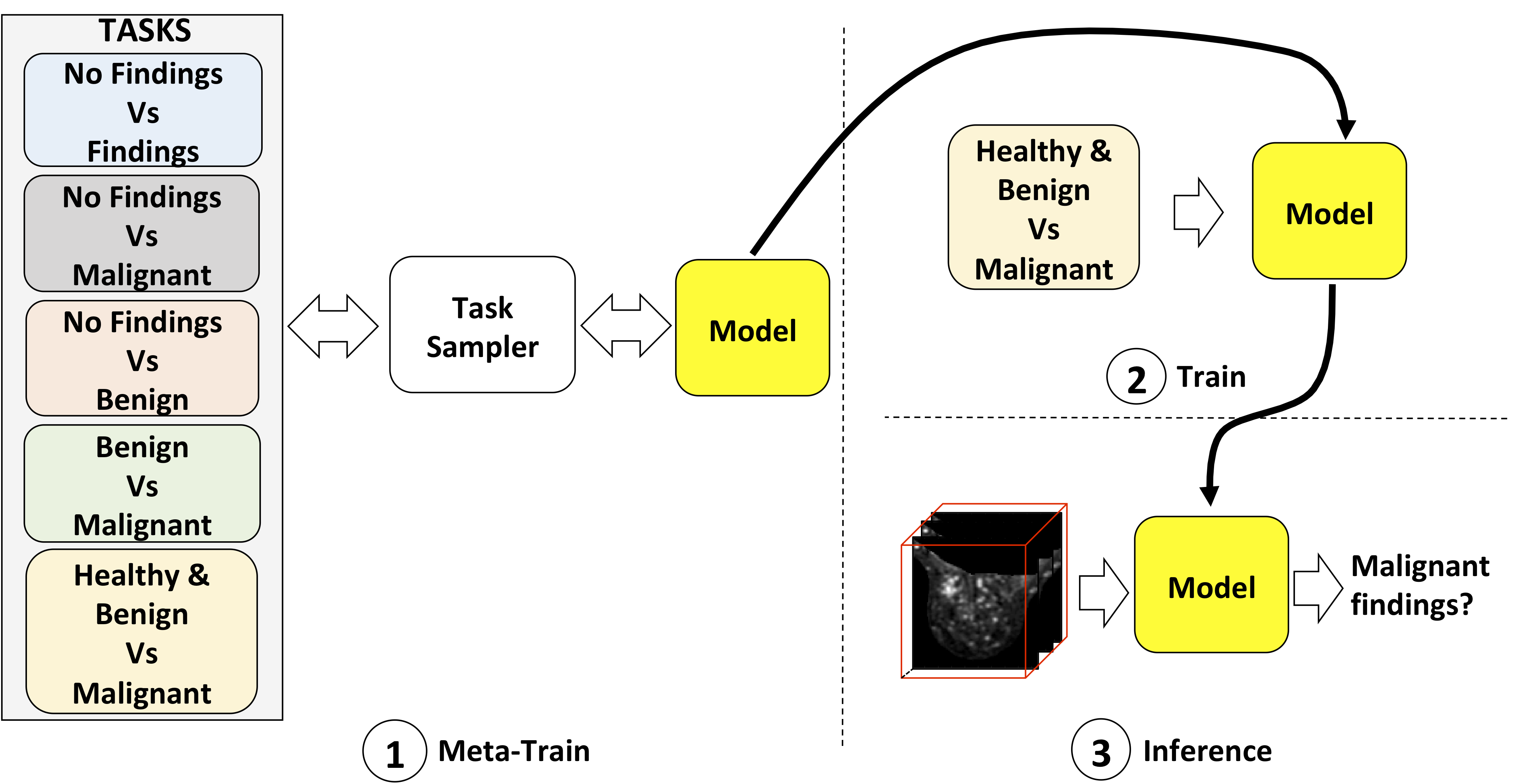} \\
\end{tabular}
\end{center}
\caption{The model is first meta-trained using several tasks containing relatively small training sets.  The meta-trained model is then used to initialize the usual training process for breast screening (i.e., healthy and benign versus malignant). The probability of malignancy is estimated from a forward pass during the inference process.}
\label{fig:intro}
\end{figure}

\vspace{-.1in}  
\section{Methodology}
\vspace{-.1in}  
\label{sec:methodology}

Our methodology consists of three stages (see Fig.~\ref{fig:intro}). We first \textbf{meta-train} the model using different tasks (each containing relatively small training sets) to find a good initialization that is then used to \textbf{train} the model for the breast screening task (i.e., the healthy and benign versus malignant task). The \textbf{inference} is performed using previously unseen test data.
Below, we define the dataset and describe each stage.

\vspace{-.1in}  
\subsection{Dataset}
\label{sec:dataMethod}
\vspace{-.05in}  

Let the dataset be represented by ${\cal D} = \left \{ \left ( \mathbf{v}_i, \mathbf{t}_i, b_i, d_i, y_i \right ) \right \}_{i=1}^{|{\cal D}|}$
where $\mathbf{v}:\Omega \rightarrow \mathbb R$ is the first subtraction DCE-MRI volume ($\Omega$ denotes the volume lattice), $\mathbf{t}:\Omega \rightarrow \mathbb R$ is the T1-weighted volume, 
$b \in \{ \text{left},\text{right} \}$ indicates if this is the left or right breast of the patient, $d_i \in \mathbb N$ denotes patient identification, and $y \in \mathcal{Y} = \{0,1,2\}$ is the volume label ($y_i = 2$: breast contains a malignant lesion, $y_i = 1$ : breast contains at least one benign and no malignant findings, and $y_i = 0$ : no findings).
We divide ${\cal D}$ using the patient identification into the training set $\mathcal{T}$, validation set $\mathcal{V}$ and testing set $\mathcal{S}$, with no overlap between these sets.
 
For the meta-training phase, we use the meta-training set defined by $\{ \mathcal{D}_j \}_{j=1}^5$ where each meta-set $\mathcal{D}_j \subseteq \mathcal{T}$ contains the relevant volumes for the classification task $K_j$, defined as follows: 
1) $K_1$ classifies volumes that contain any findings (benign or malignant); 
2) $K_2$ discriminates between volumes with no findings and malignant findings;
3) $K_3$ discriminates between volumes with no findings and benign findings; 
4) $K_4$ discriminates volumes with benign findings against  malignant findings; and
5) $K_5$ addresses breast screening, i.e. finding volumes that contain malignant findings.

\vspace{-.1in}  
\subsection{Model}
\vspace{-.05in}  
\label{sec:model}

We {\bf meta-train} a model across a number of tasks so that it can  be quickly trained to new unseen tasks from few images, or fine-tuned to become more effective at one of the tasks used in the meta-training phase. See algorithm~\ref{alg:mt} for an overview of the methodology.

\begin{algorithm}
\caption{Overview of the meta-training procedure}\label{alg:metatraining}
\begin{algorithmic}[1]
\footnotesize{
\Procedure{Meta-train}{$\{K_1 \ldots K_5\}$, $\{\mathcal{D}_1 \ldots \mathcal{D}_5\}$, model $f_{\theta}$}
\State Initialise model parameters $\theta$
	\For{$m=1\;\; to \;\; M$} \Comment{Meta-update Loop}
    \State \textbf{Create} meta-batch $\mathcal{K}_{m}$ by sampling $|\mathcal{K}_{m}|$ tasks from $\{K_1 \ldots K_5\}$ 
        \For{each task $K_j \in \mathcal{K}_m$}
		\State \textbf{Adapt} model with (\ref{eq:innerUpdt}) using samples from $\mathcal{D}_j$ \Comment{Adaptation}
		\EndFor\label{endEveryTask}
	\State \textbf{Update} model parameters with (\ref{eq:metaUpdt})\Comment{Meta-update}
	\EndFor\label{endMetaUpdate}
\EndProcedure
}
\end{algorithmic}
\label{alg:mt}
\end{algorithm}

Let $f_\theta$ be the model parameterized by $\theta$. For each meta update, the model adapts to the multiple tasks using the meta-batch set $\mathcal{K}_m$.  The tasks included in $\mathcal{K}_m$ are sampled according to one of the methods described below in Sec.~\ref{sec:taskSel}. For each task $K_j \in \mathcal{K}_m$, we sample from $\mathcal{D}_j$ a training set $\mathcal{D}_j^{tr}$ and a validation set $\mathcal{D}_j^{val}$ with $N^{tr}$ and $N^{val}$ volumes, respectively.
The model parameter $\theta$ \textbf{adaptation} is performed with the following gradient descent at time step $t$:
\begin{equation}
\theta_{j}^{\prime(t)} = \theta^{(t)} - 
\alpha \frac{\partial \mathcal{L}_{K_j} \left( f_{\theta^{(t)}} \left ( \mathcal{D}_j^{tr} \right) \right) }{\partial \theta},
\label{eq:innerUpdt}
\end{equation}
where $\alpha$ denotes the adaptation learning rate, and $\mathcal{L}_{K_j} \left( f_{\theta} \left ( \mathcal{D}_j^{tr} \right) \right)$ is the cross-entropy loss to train for the classification task $K_j$.
Finally, given the adapted models $f_{\theta_j^{\prime(t)}}$ for each task $K_j \in \mathcal{K}_m$, the model parameter $\theta$ is \textbf{meta-updated} from the error on the validation volumes $\mathcal{D}_j^{val}$ of the task w.r.t. the initial parameters $\theta^{(t)}$:
\begin{equation}
\theta^{(t+1)} = \theta^{(t)} - \beta  \mathlarger{\sum_{K_j \in \mathcal{K}_m}} \frac{ \partial \mathcal{L}_{K_j} \left( 
f_{\theta_j^{\prime(t)}}\left( \mathcal{D}_j^{val} \right)\right )}{\partial \theta},
\label{eq:metaUpdt}
\end{equation}
where $\beta$ denotes the meta-learning rate.  In summary, the {\bf meta-training} phase consists of updating the parameters of the model based on the error in validation images after being \textit{adapted} to a task using few images. This is equivalent to the following optimization:
\begin{equation}
\min_\theta \sum_{K_j \in \mathcal{K}_m} \mathcal{L}_{K_j} f_{\theta_j^{\prime(t)}}( \mathcal{D}_j^{val} ) = 
\min_\theta \sum_{K_j \in \mathcal{K}_m} \mathcal{L}_{K_j} \left(f_{\theta^{(t)} - \alpha
\frac{\partial \mathcal{L}_{K_j} \left( f_{\theta^{(t)}} \left ( \mathcal{D}_j^{tr} \right) \right)}{\partial \theta}} 
( \mathcal{D}_j^{val} )\right)\
\label{eq:metaObj}
\end{equation}

The resulting model $f_\theta$ obtained after the completion of the meta-training process is then fine-tuned using the cross entropy loss for the breast screening binary classification problem.  This process consists of the \textbf{training phase}, where we use the training set $\mathcal{T}$ for training and validation set $\mathcal{V}$ for model selection. 
The final model is tested during the \textbf{inference phase} by feeding testing volumes from $\mathcal{S}$ through the network to estimate their probability of malignancy.

\vspace{-.1in}  
\subsection{Task Sampling}
\label{sec:taskSel}

The sampling process to select $|\mathcal{K}|$ tasks from $\bigcup_{j=1}^5 K_j$ (step 4 of Alg.~\ref{alg:metatraining}) 
is currently based on random sampling~\cite{finn2017model}.  However, we consider this to be a crucial step in that algorithm, and therefore propose four sampling methods for step 4 of Alg.~\ref{alg:metatraining}.
In particular, we study the following sampling methods: 1) \textbf{Random:} randomly sample all tasks with replacement~\cite{finn2017model}; 2) \textbf{All-task:} sample all $|\mathcal{K}|=5$ tasks exactly once; 3) \textbf{Teacher-Student Curriculum Learning (CL)}~\cite{matiisen2017teacher}: sample tasks that  can achieve a higher improvement on their performance. 
This is formalized by a partially observable Markov decision process (POMDP) parametrized by 
the \textit{state}, which is the current parameter vector $\theta^{(t)}$;
the next \textit{action} to perform, which is the task $K_j$ to train on; the \textit{observation} $O_{K_j}$, consisting of the AUC improvement after adapting the parameters from $\theta^{(t)}$ to $\theta^{\prime(t)}$ for task $K_j$;
and the \textit{reward} $R_{K_j}$, which is computed from the AUC improvement of the current observation $O_{K_j}$ minus the AUC improvement obtained from the last time the task $K_j$ was sampled.
The goal of the sampling algorithm is to maximize the score of all tasks, which is solved based on reinforcement learning using Thompson sampling. 
More specifically, a buffer $\mathcal{B}_j$ stores the last $B$ rewards for task $K_j$, and at sampling time, a \textit{recent reward} is randomly chosen from each of the buffers $\mathcal{B}_j$. The next task for the meta-training is the one associated with the buffer that produced the highest absolute valued \textit{recent reward}.  This procedure chooses to lean a task until its improvement stabilizes, and then different tasks will be sampled and so on. Note that by sampling according to the absolute value, tasks where the performance is decreasing will tend to be sampled again; and 4) \textbf{Multi-armed bandit (MAB)}~\cite{gutierrez2017multi}: sample in the same way as the CL approach above, but the observation $O_{K_j}$ is stored in the buffer instead of the reward $R_{K_j}$. 
Also, the next task is selected based on the highest valued \textit{recent observation} (not its absolute value).

\vspace{-.1in}
\section{Experiments and Results}
\label{sec:Experiments} 
\vspace{-.1in}

We assess our methodology on a breast DCE-MRI dataset containing 117 patients, divided into a training set with 45 patients, a validation with 13 and a test set with 59 patients~\cite{mcclymont2014fully,maicas2017deep}.
Each sample for each patient in this dataset contains T1-weighted and dynamically-contrast enhanced MRI volumes.
Given the current interest in decreasing the number of scans~\cite{Ufuk2018Fully,maicas2017deep}, only the first subtraction volume is used.
Although all patients contain at least one lesion (benign or malignant, confirmed by biopsy), not all breasts contain lesions. 
The T1-weighted volume is only used to automatically segment and extract the left and right breasts into volumes of size  $100\times 100\times 50$~\cite{maicas2017deep} and assign separate labels to them, where the label of a breast can be "no-finding", "malignant" (if it contains at least one malignant lesion), or "benign" (if all lesions are benign).
All evaluations below are based on the area under the ROC curve (AUC).

 \begin{table}[t]
        \begin{minipage}{0.35\textwidth}
            \centering
            \vspace{.14in}
            \small{
            \begin{tabular}{|l|c|}
\hline
\textbf{Baseline} & \textbf{AUC}                                  \\
\hline
DenseNet~\cite{huang2017densely}          & 0.83                                           \\
\hline
MIL~\cite{zhu2017deep}               & 0.85                                      \\
\hline
Multi-task~\cite{xue2018full}         & 0.85                    \\
\hline
                
\end{tabular}
}
\caption{Baseline AUC \\ for classifiers trained on \\ breast screening.}
\label{tab:baselines}            
        \end{minipage}
        \hfillx
        \begin{minipage}{0.65\textwidth}
            \centering
            \small{
            \begin{tabular}{lc|c|c|c|c|}
\cline{3-6}
                                     &              & \multicolumn{4}{c|}{\textbf{AUC per sampling method}}                                      \\ \hline
\multicolumn{1}{|l|}{\textbf{Model}} & $|\mathcal{K}|$ & \textbf{Random} & \textbf{All-task} & \textbf{MAB} & \textbf{CL}   \\ \hline
\multicolumn{1}{|l|}{BSML}           & 3            & 0.86            & N/A                & 0.88            & \textbf{0.90} \\ \hline
\multicolumn{1}{|l|}{BSML}           & 5            & 0.85            & 0.89               & 0.89            & \textbf{0.90} \\ \hline
\multicolumn{1}{|l|}{BSML-NS}        & 4            & 0.85            & 0.88               & 0.87            & 0.89          \\ \hline
\end{tabular}
}
\caption{AUC for our proposed models depending on the meta-batch size and task sampling methods and trained for breast screening.}
\label{BSML_AUC}
\label{tab:bsml}
        \end{minipage}
    \end{table}


The model $f_\theta$, implemented in 3D, is based on the DenseNet~\cite{huang2017densely}, which currently holds the best classification performance in several computer vision applications.  The model architecture and hyper-parameters are selected based on the highest AUC for the breast screening problem in the validation set. The architecture is composed of five dense blocks of two dense layers each and is trained with a learning rate of 0.01 and a batch size of 2 volumes.
For our proposed methodology (labeled as BSML), the number of meta-updates is $M=3000$, the meta-learning rate $\beta = 0.001$, the number of training and validation volumes selected for task $K_j$ from the meta-set $\mathcal{D}_j$ is $N^{tr} = N^{val} = 4$, the number of gradient descent updates is $5$, and the adaptation learning rate $\alpha=0.1$.
We check the influence of the meta-batch size $|\mathcal{K}| \in \{3,5\}$. Also, we evaluate the influence of all task sampling approaches listed in Sec.~\ref{sec:taskSel}.
Finally, we also run experiments to check the performance of our model when the task of breast screening is not used for meta-training (BSML-NS).  This means that the training process has to learn an unseen task starting from the initialization achieved in the meta-training step. In this case, we use $|\mathcal{K}|=4$ and test the influence of the different task sampling approaches.

Our proposed model is compared against the following baselines:
1) a \textit{DenseNet} trained for the breast screening binary task;
2) the pre-trained DenseNet (1) fine-tuned using a multiple-instance learning framework\textit{(MIL)}~\cite{zhu2017deep} -- this approach holds the SOTA for the breast screening problem in mammography; and 
3) a DenseNet trained with a \textit{multi-task} loss~\cite{xue2018full} using the 5 tasks defined in sec.~\ref{sec:dataMethod}.

\begin{figure}[t]
\begin{tabular}{cccc}
\subfloat[]{\includegraphics[width = 1.03in]{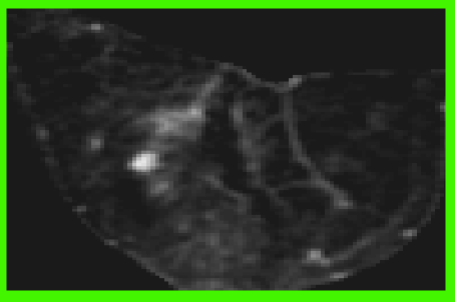}\label{img1}} &
\hspace{0.1in}
\subfloat[]{\includegraphics[width = 1.03in]{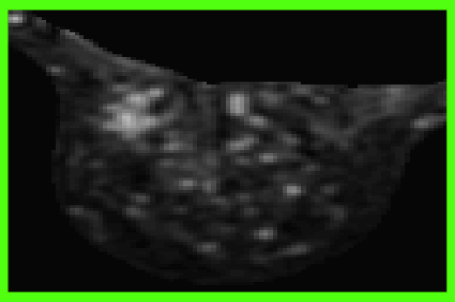}\label{img3}}&
\hspace{0.1in}
\subfloat[]{\includegraphics[width = 1.03in]{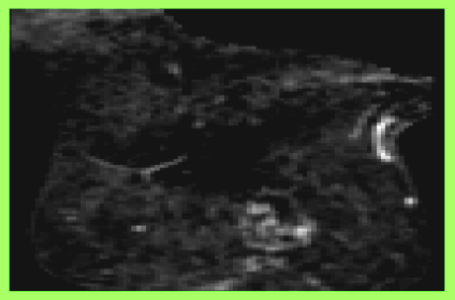}\label{img4}}&
\hspace{0.1in}
\subfloat[]{\includegraphics[width = 1.03in]{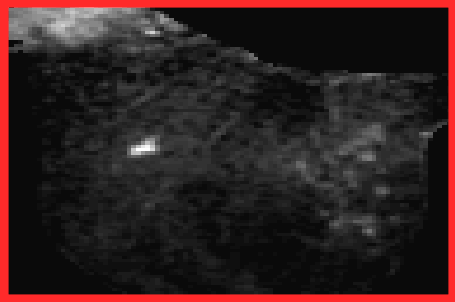}\label{img2}}
\end{tabular}
\caption{Classification examples.
Image (\ref{img1}) shows a correct negative classification of a volume containing a benign lesion,
images (\ref{img3}) and (\ref{img4}) show a correct positive classification of a volume containing a malignant lesion, and
image (\ref{img2}) shows an false negative classification of a volume containing a small malignant lesion.}
\label{fig:res}
\end{figure}

Tables~\ref{tab:baselines} and~\ref{tab:bsml} contain the AUC for baselines and experiments detailed above. Figure~\ref{fig:res} shows examples of the classification produced by our methodology.

\vspace{-.05in}
 \section{Discussion and Conclusion}
\label{sec:dic}
\vspace{-.1in}

We presented a methodology to train medical image analysis systems that tries to mimic the process of training a radiologist. 
This is achieved by meta-training the model with several tasks containing small meta-training sets, followed by a subsequent training to solve the particular problem of interest.
We established a new SOTA for the weakly supervised breast screening problem when compared to several baselines such as DenseNet~\cite{huang2017densely}, a multi-task trained DenseNet~\cite{xue2018full} and a DenseNet fine-tunned in a MIL framework~\cite{zhu2017deep}. Note that the MIL setup does not achieve a large improvement as reported in the original paper~\cite{zhu2017deep}. We believe that this is due to the use of DenseNet, which tends to show better classification results than Alexnet~\cite{zhu2017deep}.
Also, it is worth mentioning that our proposed method has not shown  any false positive classification in the test set.

As reflected in the experiments, the sampling of the tasks to meta-train is an important step of our proposed methodology. In particular, the CL sampling showed more accurate classification than random sampling, which yields similar results to the baselines. The MAB sampling improved over random selection, but it is still not as competitive as curriculum learning. We conjecture that sampling according to the best performance (i.e., MAB) keeps selecting more often the tasks that produce the highest reward, while CL samples tasks with a larger margin for improvement because they can achieve a larger slope in the learning curve. Consequently, CL aims at improving the reward for ALL tasks.
Also, the meta-batch size does not appear to have much influence in the results.
Furthermore, the BLML-NS results in Tab.~\ref{tab:bsml} show that our proposed methodology can be successfully trained for breast screening even when this task is not included in the meta-training phase.  In particular, notice that the AUC is competitive, being 1 point smaller than our best result (that includes breast screening in meta-training), but between 4 and 6 points better than the baselines. 

We would like to thank Nvidia for the donation of a TitanXp that supported this work.

\bibliographystyle{splncs}
\bibliography{biblio}
\end{document}